\pdfoutput=1

\documentclass[11pt]{article}

\usepackage[final]{acl}

\usepackage{times}
\usepackage{latexsym}

\usepackage{pgfplots}
\usepackage{longtable}
\usepackage{array} 
\usepackage{amsmath}
\PassOptionsToPackage{rgb}{xcolor}

\usepackage{tikz}
\usetikzlibrary{decorations.pathmorphing, patterns, decorations.shapes}
\pgfplotsset{compat=1.18}

\usepackage[T1]{fontenc}

\usepackage[utf8]{inputenc}

\usepackage{microtype}

\usepackage{inconsolata}

\usepackage{graphicx}
\usepackage{tabu}                     %
\usepackage{multirow}                 %
\usepackage{multicol}                 %
\usepackage{multirow}                %
\usepackage{float}                    %
\usepackage{makecell}                 %
\usepackage{booktabs}                 %
\usepackage{xspace}

\newcommand{\model}[0]{{$\mathcal{S}^2$IT}\xspace}

\title{$\mathcal{S}^2$IT: Stepwise Syntax Integration Tuning for Large Language Models in Aspect Sentiment Quad Prediction}

\author{
    Bingfeng Chen$^{1,2}$, Chenjie Qiu$^{1}$, Yifeng Xie$^{1}$, Boyan Xu$^{1}$\thanks{Corresponding author, \href{mailto:hpakyim@gmail.com}{hpakyim@gmail.com}}, Ruichu Cai$^{1,3}$, Zhifeng Hao$^{1,4}$ \\
    $^{1}$School of Computer Science, Guangdong University of Technology \\
    $^{2}$Guangdong Laboratory of Artificial Intelligence and Digital Economy (SZ) \\
    $^{3}$Peng Cheng Laboratory \\
    $^{4}$College of Science, Shantou University \\
    \texttt{chenbf@gdut.edu.cn} \\
    \texttt{\{yausankit, evfxie, hpakyim, cairuichu\}@gmail.com} \\
    \texttt{haozhifeng@stu.edu.cn} \\
}

\begin{document}
\maketitle
\begin{abstract}
Aspect Sentiment Quad Prediction (ASQP) has seen significant advancements, largely driven by the powerful semantic understanding and generative capabilities of large language models (LLMs). However, while syntactic structure information has been proven effective in previous extractive paradigms, it remains underutilized in the generative paradigm of LLMs due to their limited reasoning capabilities. In this paper, we propose $\mathcal{S}^2$IT, a novel \underline{S}tepwise \underline{S}yntax \underline{I}ntegration \underline{T}uning framework that progressively integrates syntactic structure knowledge into LLMs through a multi-step tuning process. The training process is divided into three steps. $S^2$IT decomposes the quadruple generation task into two stages: 1) Global Syntax-guided Extraction and 2) Local Syntax-guided Classification, integrating both global and local syntactic structure information. Finally, \emph{Fine-grained Structural Tuning} enhances the model’s understanding of syntactic structures through the prediction of element links and node classification. Experiments demonstrate that $\mathcal{S}^2$IT significantly improves state-of-the-art performance across multiple datasets. Our implementation will be open-sourced at
\url{https://github.com/DMIRLAB-Group/S2IT}.

\end{abstract}

\begin{figure}[htb]
    \centering
    \includegraphics[width=1\linewidth]{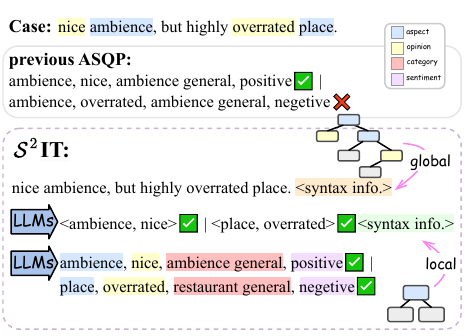}
  \caption{An example of the ASQP task. The most notable feature that distinguishes $\mathcal{S}^{2}$IT from previous work is that it decomposes the quadruples and injects syntactic information from different perspectives step by step into the large language model.}
  \label{fig1}
\end{figure}

\begin{figure*}[htb]
    \centering
    \includegraphics[width=1\linewidth]{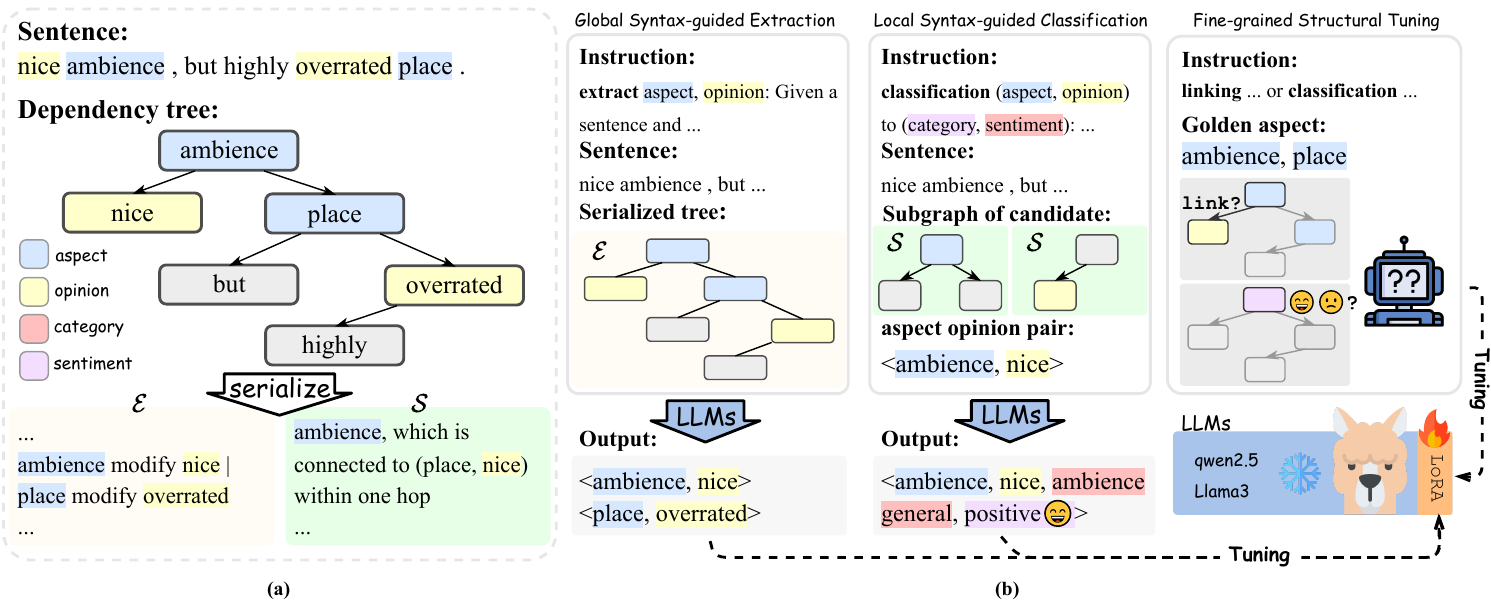}
    \caption{$\mathcal{S}^2$IT framework illustration. (a) shows how we serialize the dependency tree into natural language. (b) shows the main part of our framework and two structure instruction tuning tasks.}
    \label{fig2}
\end{figure*}

\section{Introduction}
Aspect Sentiment Quad Prediction (ASQP) focuses on predicting tuples of sentiment-related elements from a given text \citep{zhang2022surveyaspectbasedsentimentanalysis}. These elements consist of four components central to ASQP: aspect term (a), aspect category (c), opinion term (o), and sentiment polarity (s). For example, in the sentence "\textit{I really love sushi!}", the corresponding elements are "\textit{sushi}" (aspect term), "\textit{food quality}" (aspect category), "\textit{love}" (opinion term), and "\textit{positive}" (sentiment polarity). Although ASQP is fundamentally an extractive task, the ability to generalize and generate responses has made fine-tuning models like T5~\cite{raffel2023exploringlimitstransferlearning} a mainstream approach, as demonstrated by \citet{zhang-etal-2021-aspect-sentiment}.

Recent approaches to improving language models' performance have incorporated syntactic structure knowledge, typically divided into encoder-only and encoder-decoder. For instance, encoder-only approaches \citep{liang-etal-2022-bisyn,li-etal-2023-dual-channel,chen2024s2gslincorporatingsegmentsyntactic} enhance the connections between elements at the representation level through Graph Neural Networks (GNNs).
For encoder-decoder approaches, a key challenge is how to add syntactic structure knowledge to language models.
\citet{li2023graphix} and \citet{SynGen} incorporates GNNs into the encoder layer of T5 and BART. 
These approaches are limited by their reliance on GNN integration, which does not extend well to decoder-only language models.

In this paper, we propose $\mathcal{S}^2$IT, which introduces a novel \underline{S}tepwise \underline{S}yntax \underline{I}ntegration \underline{T}uning framework. Specifically, we decompose the complex task of generating sentiment quadruples into two stages: \emph{Global Syntax-guided Extraction} and \emph{Local Syntax-guided Classification}. 
First, \emph{Global Syntax-guided Extraction} incorporates global syntactic knowledge to identify aspects and opinions. Second, \emph{Local Syntax-guided Classification} classifies opinions into sentiment and category by utilizing the local syntactic relationships between aspects and opinions. As shown in Figure \ref{fig1}, the incorporation of syntactic information effectively helps $\mathcal{S}^2$IT mitigate the influence of multi-aspect terms.

We also introduce \emph{Fine-grained Structural Tuning } to improve LLMs' ability to understand and use structural knowledge for reasoning. 
Our main contributions are as follows:
(1) We propose $\mathcal{S}^2$IT, a novel Stepwise Syntax Integration Tuning framework that progressively integrates syntactic structure knowledge into large language models (LLMs) through a multi-step tuning process.
(2) $\mathcal{S}^2$IT effectively learns sentiment element relationships by leveraging syntactic structure at different granularities and local levels through three stages: Global Syntax-guided Extraction, Local Syntax-guided Classification, and Fine-grained Structural Tuning.
(3) Experiments demonstrate that $\mathcal{S}^2$IT achieves the state-of-the-art on the restaurant and laptop datasets.

\section{Methodology}
We first formulate the task of ASQP (\S\ref{tab:problem}). As shown in Figure~\ref{fig2}, we then present our \model, which was enhanced with the syntax information. It includes the extraction task (\S\ref{tab:extraction}), classification task (\S\ref{tab:classfication}), and structure instruction tuning (\S\ref{tab:tuning}).

\subsection{Problem Formulation} \label{tab:problem}

For a given sentence $x=\{x_{1},x_{2},...,x_{n}\}$, ASQP aims to predict all the aspect-level sentiment quadruples $\{(a_{j}, o_{j}, c_{j}, s_{j})\}^{N}_{j=1}$, which represent the \textbf{a}spect term, \textbf{o}pinion term, \textbf{c}ategory term and \textbf{s}entiment polarity, respectively. The aspect category $c$ belongs to a predefined set of categories, and the sentiment polarity $s$ falls into one of three categories: negative, positive, and neutral.

\subsection{Global Syntax-guided Extraction} \label{tab:extraction}
Although promising, current methods fall short of fully leveraging the additional syntactic features available. Our objective is to extract potential aspect-opinion pairs $\mathcal{P}=\{\langle a_j, o_j \rangle\}^{N}_{j=1}$ from sentences, We can introduce syntactic information to help the model discover relationships between the elements.

As an example, we consider the injection of modifier relationships between elements into LLMs. We utilize the template $\mathcal{E}$ to delineate the type of relation found in the dependency tree. Assumed that $x_{i}$ serves as the head node for $x_{h}$, we document the details of this syntactic structure as follows:

\begin{equation}
    \mathcal{E}_{i\in\{1,...,n\}}= \{x_{i}\}\ \mathrm{modify}\ \{x_{h}\}
\end{equation}

Afterward, we concatenate all pairs according to the order of their occurrence within the sentence to create a natural language description of the global structure. This description is subsequently appended to the original sentence, enhancing the syntactic context.
Furthermore, we implement instruction tuning using the instruction $\mathcal{I}_1$ and the output $\mathcal{O}_1$. 
The process is as follows: 
\begin{equation}
  \mathcal{O}_{1,j\in\{1,...,N\}} = \mathrm{aspect}\colon\{a_{j}\},\mathrm{opinion} \colon\{o_{j}\}
\end{equation}
The objective of supervised fine-tuning is to minimize the loss defined as:
\begin{equation}
    \min_{M^{*}}\mathcal{L}(M^{*}(\mathcal{I}_1,\ x,\ \mathcal{E}_1|...|\mathcal{E}_n,\mathcal{O}_{1,1}|...|\mathcal{O}_{1,N}))
\end{equation}

\subsection{Local Syntax-guided Classification } \label{tab:classfication}
Focusing not only on the global syntactic structure, which has been shown to enhance performance~\cite{cheng-etal-2023-mrrl,xie-etal-2023-syntax}, but also paying attention to the local syntactic structure is crucial~\cite{liang-etal-2022-bisyn,chen2024s2gslincorporatingsegmentsyntactic}. After obtaining the correct pairs $\mathcal{P}$, treating these pairs as central nodes in the dependency tree, we extract neighboring words as subgraphs. For example, the description of a subgraph $\mathcal{S}$ centered on $a_{j}$, including at most one-hop neighbor details, is:

\begin{equation}
    \label{t2}
    \begin{split}
        \mathcal{S}^{(1)}_{a_{j\in\{1,...,N\}}} = &\{a_{j}\} \ \mathrm{is\ connected\ with\ }\\
        &\{ v \mid \mathcal{A}_{a_{j}v} = 1 \} \ \mathrm{within\ one\ hop.}
    \end{split}
\end{equation}
where $\mathcal{A}$ represents the adjacency matrix and $\{ v \mid \mathcal{A}_{a_{j}v} = 1 \}$ denotes all one-hop neighbors of $a_j$ in the dependency tree. The formula also holds true for $o_j$ and $\mathcal{S}^{(1)}_{o_{j}}$.

Subsequently, we also concatenate this local syntactic information with the original sentence, integrating the output $\mathcal{O}_{2}$ and the instruction $\mathcal{I}_2$. This integrated data is then processed and tuned to optimize model performance further:
\begin{equation}
    \begin{split}
  \mathcal{O}_{2,j\in}&_{\{1,...,N\}} = \mathrm{aspect}\colon\{a_{j}\},\mathrm{opinion} \colon\{o_{j}\} \\
                        & \mathrm{category}\colon\{c_j\},\mathrm{sentiment}\colon\{s_j\}
    \end{split}
\end{equation}
We then fine-tune the LLMs while minimizing the following loss function:

\begin{equation}
    \begin{split}
        \min_{M^{*}} \mathcal{L}( M^{*}(\mathcal{I}_2, &x, \mathcal{S}^{(1)}_{a_1}, \mathcal{S}^{(1)}_{o_1} \mid \ldots \mid \mathcal{S}^{(1)}_{a_N}, \mathcal{S}^{(1)}_{o_N},\mathcal{P}, \\
        &\mathcal{O}_{2,1} \mid \ldots \mid \mathcal{O}_{2,N}))
    \end{split}
\end{equation}

\subsection{Fine-grained Structure Instruction Tuning} \label{tab:tuning}

\paragraph{Element Link Prediction.} 
To tailor our model to specialized structured knowledge tasks, we introduce the auxiliary task about element link prediction. Specifically, we developed a structure-aware matching task that presents genuine sentiment elements (e.g., aspects) to LLMs and uses their corresponding sentiment elements (e.g., opinions) as labels. This approach guides the model to accurately associate them based on the provided sentences and syntactic information. Such alignment not only enhances the model's accuracy but also its capability to manage complex relationships within the data.

\paragraph{Node Classification.} 
Building upon our method, we introduce an auxiliary task focused on sentiment element classification. This task is integral to our structure-aware strategy, directing the model to accurately assign labels to each sentiment element (e.g., sentiment polarity or other relevant categories) within the structural framework. Through this task, the model effectively learns semantic information and local syntactic structures, thereby enhancing its capability to comprehend and analyze intricate language nuances.

These tasks are integral to our fine-grained structure instruction tuning method.
All our instruction prompts are summarized in Appendix~\ref{Appendix: Tuning}.
\section{Experiments}

\subsection{Datasets}

We validate our methods on Restaurant and Laptop datasets \citep{acos}. There are 2,286 sentences in Restaurant domain, and 4,076 sentences in Laptop domain. Following the setting from \citet{acos}, we divide the original dataset into a training set, a validation set, and a testing set.

\subsection{Implement Details}

We used Qwen2.5-7B-Instruct\footnote{\url{https://huggingface.co/Qwen/Qwen2.5-7B-Instruct}} and Qwen2.5-32B-Instruct\footnote{\url{https://huggingface.co/Qwen/Qwen2.5-32B-Instruct}} as the base language model. In addition, to verify how well our framework works on different LLMs, we also conducted ablation experiments on Llama3-8B-instruct\footnote{\url{https://huggingface.co/meta-llama/Meta-Llama-3-8B-Instruct}} in (\S\ref{tab:Structural Tuning}). 
The model was trained with an initial learning rate of 5e-5 for 5 epochs. We set the batch size to 4 and used a gradient accumulation of 2. We applied LoRA \cite{hu2021loralowrankadaptationlarge} with a LoRA rank of 32 for efficient fine-tuning, allowing our 7B model to be trained and inferred on a single NVIDIA GeForce RTX 3090 24G. All training details, including hyperparameters, will be presented in Appendix~\ref{Appendix: Hyper}.

\subsection{Main results}
\begin{table}
    \centering
    \Large
    \renewcommand\arraystretch{1.3}
    \scalebox{0.6}{
        \begin{tabular}{lcccc}
            \toprule
            \multirow{2}{*}{\textbf{Model}} & \multicolumn{2}{c}{Restaurant} & \multicolumn{2}{c}{Laptop} \\
            \cmidrule(lr){2-3}\cmidrule(lr){4-5}
            & R & F1 & R & F1 \\
            \midrule
            $\bullet$ \textbf{\textit{Extractive-based methods}} & & & & \\
            TAS-BERT\citep{Wan_Yang_Du_Liu_Qi_Pan_2020}           & 46.3 & 33.5 & 19.2 & 27.3 \\
            Extract-Classify\citep{cai2021aspect}                 & 52.9 & 44.6 & 29.4 & 35.8 \\
            One-ASQP\citep{zhou2023unifiedonestepsolutionaspect}  & 56.2 & 60.6 & 39.5 & 41.5 \\
            \midrule
            $\bullet$ \textbf{\textit{Generative-based methods}} & & & & \\
            Paraphrase\citep{zhang-etal-2021-aspect-sentiment}    & 59.8 & 59.8 & 42.5 & 43.0 \\
            DLO\citep{hu-etal-2022-improving-aspect}              & 59.8 & 59.9 & 43.8 & 43.6 \\
            \quad+AToss\citep{seo2024makecompoundsentencessimple} & 59.9 & 60.5 & 43.9 & 44.5 \\
            MvP\citep{mvp}                                       & 57.8 & 59.5 & 43.6 & 43.7 \\
            \quad+AToss\citep{seo2024makecompoundsentencessimple} & 58.3 & 60.6 & 43.1 & 44.2 \\
            \midrule[0.08em]
            qwen2.5-7B                                            & 64.1 & 63.6 & 44.7 & 44.6 \\
            \textbf{$\mathcal{S}^2$IT-7B}                        & 64.9 & \underline{66.1} & 44.7 & \underline{45.9} \\
            \midrule
            qwen2.5-32B                                           & \underline{65.0} & 64.2 & \underline{45.0} & 45.3 \\
            \textbf{$\mathcal{S}^2$IT-32B}                       & \textbf{66.6} & \textbf{67.4} & \textbf{45.4} & \textbf{46.7} \\
            \bottomrule
        \end{tabular}
    }
    \caption{Recall and F1 scores (\%) on the Restaurant and Laptop datasets compared to other baselines.}
    \label{table1}
\end{table}

Our method significantly outperforms previous state-of-the-art baselines in terms of F1 scores across five datasets in supervised settings as shown in Table~\ref{table1}, becoming SOTA for all these tasks. In particular, our performance on the restaurant dataset of the challenging ACOS task exceeds the baseline model by 2.71 points. This improvement is attributed to our two-step framework, which effectively reduces the coupling in the tuple generation task, helping the model extract correct aspect-opinion pairs from the sentence and make accurate classifications from both syntactic and semantic perspectives.

\subsection{Ablation Experiment.}
\begin{figure}[ht]
    \centering
    \begin{tikzpicture}
        \begin{axis}[
            width=4.5cm,  %
            height=3.5cm,  %
            ybar,  %
            bar width=10pt,  %
            symbolic x coords={step 1, step 2},  %
            xtick=data,  %
            xticklabel style={yshift=3pt,draw=none},  %
            xlabel={RESTAURANT},  %
            ylabel={F1 score},  %
            ytick={70, 80, 90},
            ymin=65,  %
            enlarge x limits=0.6,  %
            grid=major,  %
            xmajorgrids=false,
            grid style={dashed},
            tick label style={font=\scriptsize},
            xlabel style={yshift=0pt,font=\small},
            ylabel style={yshift=-15pt,font=\small},
            legend style={at={(0.03,0.55)}, 
            anchor=south west,legend columns=1, 
            font=\small, 
            nodes={scale=0.6, transform shape},
            draw=none,
            fill=none
            }  %
        ]

        \addplot [
            pattern=north east lines,  %
            pattern color=yellow!60!red,  %
            draw=yellow!60!red, %
            line width=0.5pt  %
        ] coordinates {(step 1, 72.2) (step 2, 87.8)};
        \addplot  coordinates {(step 1, 74.3) (step 2, 89.1)};
        \legend{w/o syntax, w syntax}
        \end{axis}
    \end{tikzpicture}
    \begin{tikzpicture}
        \begin{axis}[
            width=4.5cm,  %
            height=3.5cm,  %
            ybar,  %
            bar width=10pt,  %
            symbolic x coords={step 1, step 2},  %
            xtick=data,  %
            xticklabel style={yshift=3pt},  %
            xlabel={LAPTOP},  %
            ylabel={},  %
            ytick={60, 70, 80, 90},
            ymin=55,  %
            enlarge x limits=0.6,  %
            grid=major,  %
            xmajorgrids=false,
            grid style={dashed},
            tick label style={font=\scriptsize},
            xlabel style={yshift=0pt,font=\small},
            ylabel style={yshift=-12pt,font=\small},
            legend style={at={(0.5,-0.15)}, anchor=north,legend columns=-1}  %
        ]
        \addplot [
            pattern=north east lines,  %
            pattern color=yellow!60!red,  %
            draw=yellow!60!red, %
            line width=0.5pt  %
        ]  coordinates {(step 1, 71.2) (step 2, 61.0)};
        \addplot coordinates {(step 1, 73.5) (step 2, 62.4)};
        \end{axis}
    \end{tikzpicture}
    \caption{The impact of syntactic information at each stage on the Laptop and Restaurant datasets. Step 1 demonstrated the improvements that syntactic information brings to the extraction of $\mathcal{P}$, Step 2 showed how much the model depends on syntactic information when it has the correct $\mathcal{P}$.}
    \label{fig3}
\end{figure}
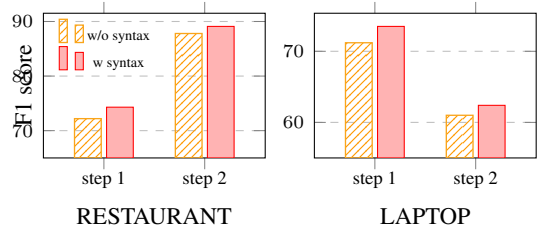

\paragraph{Effect Analysis of syntax integration.}\label{tab:Structural Templates}
To verify the impact of syntax integration, we conducted additional experiments to evaluate the effectiveness of syntax information in our framework. Table~\ref{table2} compares the base model (qwen2.5-7B), the full $\mathcal{S}^2$IT framework, and $\mathcal{S}^2$IT with syntax information entirely removed. The results highlight the importance of syntax information in achieving superior performance.
\paragraph{Effect Analysis of Syntax Imformation.}
Global structural information effectively helps the model understand the syntactic relationship between aspects and opinions in a sentence. As shown in Figure~\ref{fig3}, incorporating global syntactic features significantly improved the model's performance. We observed that, compared to the model without syntactic injection, our approach increased the F1 score by 2.9\% on the Restaurant dataset and by 3.1\% on the Laptop dataset in Step 1, proving the importance of global structure in handling complex dependencies.
Local structure helps the model capture fine-grained details by aggregating local syntactic information. In Step 2, incorporating local structure improved the F1 score by 1.4\% on the Restaurant dataset and 2.2\% on the Laptop dataset. Thus, understanding nearby sentiment elements significantly enhances the model's ability significantly to perform accurate sentiment classification.

\begin{table}[ht]
    \centering
    \Large
    \renewcommand\arraystretch{1.3}
    \scalebox{0.6}{
        \begin{tabular}{lcccc}
            \toprule %
            \multirow{2}{*}{\textbf{Model}} & \multicolumn{2}{c}{Restaurant} & \multicolumn{2}{c}{Laptop} \\ 
            \cmidrule(lr){2-3} \cmidrule(lr){4-5}
            & R & F1 & R & F1 \\
            \midrule
            qwen2.5-7B & 64.1 & 63.6 & 44.7 & 44.6 \\
            \textbf{$\mathcal{S}^2$IT-7B} & \textbf{64.9} & \textbf{66.1} & \textbf{45.0} & \textbf{45.9} \\
            \quad w/o syntax information & 63.0 & 64.8 & 43.6 & 44.8 \\
            \bottomrule
        \end{tabular}   
    }
    \caption{Full $\mathcal{S}^2$IT framework, with Fine-grained Structural Tuning, but completely excluding syntax information}
    \label{table2}
\end{table}

\begin{table}[ht]
    \centering
    \Large
    \renewcommand\arraystretch{1.3}
    \scalebox{0.6}{
        \begin{tabular}{lcccc}
            \toprule %
            \multirow{2}{*}{\textbf{Model}}& \multicolumn{2}{c}{Restaurant} & \multicolumn{2}{c}{Laptop}\\ 
            \cmidrule(lr){2-3}\cmidrule(lr){4-5}
            & R & F1 & R & F1 \\
            \midrule
            {$\mathcal{S}^2$IT+qwen2.5(7B)}                & 64.9 & 66.1 & 44.7 & 45.9 \\
            \midrule
            \quad w/o Element Link               & 63.7 & 65.3 & 43.4 & 44.7 \\
            \quad w/o Node Classification        & 63.1 & 65.0 & 43.3 & 44.5 \\
            \qquad w/o both                      & 62.7 & 64.3 & 42.9 & 44.5 \\
            \midrule
            \midrule
            $\mathcal{S}^2$IT+llama3(8B)                  & 64.8 & 66.0 & 44.2 & 45.7 \\
            \midrule
            \quad w/o Element Link               & 62.7 & 64.9 & 43.6 & 44.8 \\
            \quad w/o Node Classification        & 63.5 & 65.8 & 42.3 & 43.9 \\
            \qquad w/o both                      & 62.8 & 64.3 & 43.4 & 44.6 \\
            \bottomrule
        \end{tabular}
    }
    \caption{The ablation experiments on Structure Instruction Tuning.}
    \label{table3}
\end{table}

\paragraph{Effect Analysis of Structure Instruction Tuning.}\label{tab:Structural Tuning}
To validate the effectiveness of structural instruction tuning, we conducted ablation experiments on two datasets using two different LLMs, focusing on the impact of element linking and node classification on the model’s ability to understand structural information. As shown in Table~\ref{table3}, the effectiveness of our tasks was confirmed. We observed a significant performance drop across both models on different datasets after removing all structural instruction tuning tasks. This result indicates that the structural instruction tuning tasks effectively enhance the model's understanding of hierarchical structures, especially when handling tasks like complex sentiment classification task.

\paragraph{Effect of Structural Prompt Templates.}
To further validate our structural prompt templates, we designed an interesting experiment comparing the overall impact of dependency tree templates composed of natural language (nl-syn) versus classical dependency tree templates composed of symbols (symbol-syn) on our method. The results in Table~\ref{table4} reveal that symbol-based templates significantly degrade performance. We attribute this decline to the inconsistency between symbol templates and the pretraining objectives of LLMs, which adversely affects downstream task performance. This experiment demonstrates the rationale for converting syntactic information into natural language to better align with the capabilities of LLMs.

\begin{table}
    \centering
    \Large
    \renewcommand\arraystretch{1.3}
    \scalebox{0.6}{
        \begin{tabular}{lcccc}
            \toprule %
            \multirow{2}{*}{\textbf{Model}} & \multicolumn{2}{c}{Restaurant} & \multicolumn{2}{c}{Laptop} \\ 
            \cmidrule(lr){2-3} \cmidrule(lr){4-5}
            & R & F1 & R & F1 \\
            \midrule
            qwen2.5-7B    & 64.1 & 63.6 & 44.7 & 44.6 \\
            qwen2.5-7B  w/ \textbf{nl-syn}         & \textbf{65.3} & \textbf{64.4} & \textbf{45.0} & \textbf{44.9} \\
            qwen2.5-7B  w/ symbol-syn     & 61.9 & 61.0 & 42.8 & 42.9 \\
            \bottomrule
        \end{tabular}   
    }
    \caption{Performance Comparison of Different Dependency Tree Templates on Restaurant and Laptop Datasets. Here, nl-syn refers to dependency tree syntactic information described using natural language templates, while symbol-syn refers to dependency tree syntactic information presented in the traditional bracketed tree format.}
    \label{table4}
\end{table}

\section{Related Work}
\paragraph{Extractive Methods}Previous methods leverage language models such as BERT to generate word-level textual embeddings and train additional classification heads \cite{Wan_Yang_Du_Liu_Qi_Pan_2020}. \citet{cai2021aspect,zhou2023unifiedonestepsolutionaspect} employ a joint multi-task sequence labeling approach to extract elements within tuples. However, their specially designed extraction framework exhibits poor generalization across various ABSA tasks. Consequently, unified generative methods have become the current mainstream.
\paragraph{Generative Methods}Recent approaches have favored designing a unified end-to-end framework for performing ABSA tasks using generative language models. In essence, these methods serialize target tuples into natural language and employ them as labels to fine-tune the language models \cite{zhang-etal-2021-aspect-sentiment}. 

Depending on the task requirements, sentiment tuples can be transformed into various sequences for training. \citet{yan2021unifiedgenerativeframeworkaspectbased} use the indices of the target sentiment elements within a sentence as labels to guide the model in generating the indices of aspects and opinions. \citet{ijcai2022p561,bao2023exploring} introduced a tree generation template that guides the model in capturing the semantic relationships between sentiment elements and utilizes a generative model to extract linearized trees.

A key challenge is how to add syntactic structure knowledge to language models.
\citet{li2023graphix} and \citet{SynGen} incorporates GNNs into the encoder layer of T5 and BART. These methods all face the challenge of incorporating syntactic structural information into decoder-only models. Our framework effectively integrates syntactic knowledge into decoder-only models in a seamless and fluent manner, without requiring any structural modifications to the models.

\section{Conclusion}
In this paper, we presented $\mathcal{S}^2$IT, a novel framework designed to enhance Aspect Sentiment Quad Prediction (ASQP) by incorporating syntactic structure knowledge into large language models (LLMs) through a multi-step tuning process. The core idea of our framework is to simultaneously decompose quadruple prediction and syntactic structure learning during the multi-step tuning process of language models. The significant performance gains and adaptability to different language models further validate the effectiveness of our fine-tuning framework. 

\section*{Limitations}
While our framework achieved state-of-the-art results on the Restaurant and Laptop datasets, there is still room for improvement in constructing syntactic information and designing structural instruction tuning tasks. We have only implemented classic structure-aware tasks such as link prediction and node classification, yet these have significantly boosted performance. Nonetheless, enhancing large language models (LLMs) to better understand and reason about structural relationships remains an ongoing challenge.

\section*{Acknowledgments}
This research was financially supported by the Open Research Fund from Guangdong Laboratory of Artificial Intelligence and Digital Economy (SZ), under Grant No. GML-KF-24-23, National Science and Technology Major Project (2021ZD0111501), National Science Fund for Excellent Young Scholars (62122022), Natural Science Foundation of China (62406078, 62476163, U24A20233), the major key project of PCL (PCL2021A12), Guangdong Basic and Applied Basic Research Foundation (2023B1515120020), Collaborative Education Project of the Ministry of Education (202407). This research was enabled by the computational resources and support of the High Performance Computing Platform at the School of Computer Science, Guangdong University of Technology.

\bibliography{custom}

\section*{Appendix}
\appendix

\section{Training}

\subsection{Details of Hyperparameters} \label{Appendix: Hyper}

In this part, we outline the key hyperparameters used for fine-tuning the three different LLMs in our experiments: Qwen2.5-7B-instruct, Qwen2.5-32B-instruct and Llama3-8B-instruct. The 32B model was trained on an NVIDIA A100 80G. LoRA was applied to fine-tune the models, with ranks set to 16, or 32, and alpha values set to either 32. Dropout rates were adjusted between 0.1 to mitigate overfitting.

\begin{table}[htb]
    \centering
    \begin{tabular}{|l|l|}
        \hline
        \textbf{Hyperparameter}            & \textbf{Value}        \\ \hline
        Learning Rate                      &  5e-5                    \\ \hline
        Num Train Epochs                   &  5                    \\ \hline
        LoRA Rank                          &  32                    \\ \hline
        LoRA Alpha                         &  32                    \\ \hline
        LoRA Dropout                       &  0.1                    \\ \hline
        Batch Size                         &  4                    \\ \hline
        Gradient Accumulation Steps        &  2                    \\ \hline
        LR Scheduler Type                  &  cosine                    \\ \hline
    \end{tabular}
    \caption{Hyperparameters used in the Qwen2.5-7B-instruct.}
\end{table}
\begin{table}[htb]
    \centering
    \begin{tabular}{|l|l|}
        \hline
        \textbf{Hyperparameter}            & \textbf{Value}        \\ \hline
        Learning Rate                      &  5e-5                    \\ \hline
        Num Train Epochs                   &  5                    \\ \hline
        LoRA Rank                          &  32                    \\ \hline
        LoRA Alpha                         &  32                    \\ \hline
        LoRA Dropout                       &  0.1                    \\ \hline
        Batch Size                         &  2                   \\ \hline
        Gradient Accumulation Steps        &  4                   \\ \hline
        LR Scheduler Type                  &  cosine                    \\ \hline
    \end{tabular}
    \caption{Hyperparameters used in the Qwen2.5-32B-instruct.}
\end{table}
\begin{table}[!htb]
    \centering
    \begin{tabular}{|l|l|}
        \hline
        \textbf{Hyperparameter}            & \textbf{Value}        \\ \hline
        Learning Rate                      &  5e-5                    \\ \hline
        Num Train Epochs                   &  5                    \\ \hline
        LoRA Rank                          &  16                    \\ \hline
        LoRA Alpha                         &  32                    \\ \hline
        LoRA Dropout                       &  0.1                    \\ \hline
        Batch Size                         &  4                    \\ \hline
        Gradient Accumulation Steps        &  2                    \\ \hline
        LR Scheduler Type                  &  cosine                    \\ \hline
    \end{tabular}
    \caption{Hyperparameters used in the Llama3-8B-instruct.}
\end{table}

\subsection{Details of Instruction Tuning} \label{Appendix: Tuning}
We provide comprehensive details of the instruction tuning process. Specifically, we present examples of tasks used during Supervised Fine-Tuning (SFT). These examples illustrate the types of instructions and responses the model was trained on, highlighting the integration of domain-specific knowledge and structural information to align the model with the desired tasks.

\begin{table}[!htb]
    \centering
    \tiny
    \begin{tabular}{|p{0.45\textwidth}|}
        \hline
        \rule{0pt}{8pt}
        \centerline{\small{\textbf{task: (extract aspect, opinion)}}}\\
        \hline
        \textbf{Instruction}: \\
        Given a sentence and related dependency relations, extract aspect and opinion (both implicit and explicit) from the sentence and return pair(aspect, opinion). Pay attention to the one or multi hop dependency relationships between aspect and opinion.\\
        \textbf{Input}: \\
        sentence: service ok but unfriendly , filthy bathroom . \\ dependency relation: root depend service | service modify ok | bathroom depend but | bathroom modify unfriendly | bathroom modify filthy | service depend bathroom\\
        \textbf{Output}:\\
        aspect: service, opinion: ok | aspect: service, opinion: unfriendly | aspect: bathroom, opinion: filthy \\
        < | end\_of\_sentence | > \\

        \hline
    \end{tabular}
\end{table}
\begin{table}[!htb]
    \centering
    \tiny
    \begin{tabular}{|p{0.45\textwidth}|}
        \hline
        \rule{0pt}{8pt}
        \centerline{\small{\textbf{task: (extract opinion, aspect)}}}\\
        \hline
        \textbf{Instruction}: \\
        Given a sentence and related dependency relations, extract opinion and aspect (both implicit and explicit) from the sentence and return pair(opinion, aspect). Pay attention to the one or multi hop dependency relationships between aspect and opinion.\\
        \textbf{Input}: \\
        sentence: service ok but unfriendly , filthy bathroom . \\ dependency relation: root depend service | service modify ok | bathroom depend but | bathroom modify unfriendly | bathroom modify filthy | service depend bathroom\\
        \textbf{Output}:\\
        opinion: ok, aspect: service | opinion: unfriendly, aspect: service | opinion: filthy, aspect: bathroom \\
        < | end\_of\_sentence | > \\

        \hline
    \end{tabular}
\end{table}
\begin{table}[!htb]
    \centering
    \tiny
    \begin{tabular}{|p{0.45\textwidth}|}
        \hline
        \rule{0pt}{8pt}
        \centerline{\small{\textbf{task: (linking aspect to opinion)}}}\\
        \hline
        \textbf{Instruction}: \\
        Given a sentence, related dependency relations and known aspects, determine the opinion (both implicit and explicit) related to the each aspect from dependency relation and return the pair(aspect, opinion).\\
        \textbf{Input}: \\
        sentence: service ok but unfriendly , filthy bathroom . \\ dependency relation: root depend service | service modify ok | bathroom depend but | bathroom modify unfriendly | bathroom modify filthy | service depend bathroom\\
        candidates: aspect: service | aspect: service | aspect: bathroom \\
        \textbf{Output}:\
        aspect: service, opinion: ok | aspect: service, opinion: unfriendly | aspect: bathroom, opinion: filthy \\
        < | end\_of\_sentence | > \\

        \hline
    \end{tabular}
\end{table}
\begin{table}[!htb]
    \centering
    \tiny
    \begin{tabular}{|p{0.45\textwidth}|}
        \hline
        \rule{0pt}{8pt}
        \centerline{\small{\textbf{task: (linking opinion to aspect)}}}\\
        \hline
        \textbf{Instruction}: \\
        Given a sentence, related dependency relations and known opinions, determine the aspect (both implicit and explicit) related to the each opinion from dependency relation and return the pair(opinion, aspect).\\
        \textbf{Input}: \\
        sentence: service ok but unfriendly , filthy bathroom . \\ dependency relation: root depend service | service modify ok | bathroom depend but | bathroom modify unfriendly | bathroom modify filthy | service depend bathroom\\
        candidates: opinion: ok | opinion: unfriendly | opinion: filthy \\
        \textbf{Output}:\\
        opinion: ok, aspect: service | opinion: unfriendly, aspect: service | opinion: filthy, aspect: bathroom \\
        < | end\_of\_sentence | > \\

        \hline
    \end{tabular}
\end{table}

\begin{table}[!htb]
    \centering
    \tiny
    \begin{tabular}{|p{0.45\textwidth}|}
        \hline
        \rule{0pt}{8pt}
        \centerline{\tiny{\textbf{task: (classification (aspect, opinion) to (category, sentiment))}}}\\
        \hline
        \textbf{Instruction}: \\
        Given a sentence, related dependency relations (will be presented in the form of subgraph) and (aspect, opinion) candidates, determine the category of the aspect and the sentiment (positive, neutral, negative) of the opinion and return the quadruple(aspect, opinion, category, sentiment).\\
        \textbf{Input}: \\
        sentence: service ok but unfriendly , filthy bathroom . \\ subgraph: aspect: service, which is connected to (bathroom, ok) within one hop. opinion: ok, which is connected to (service) within one hop. | aspect: service, which is connected to (bathroom, ok) within one hop. opinion: unfriendly, which is connected to (bathroom) within one hop. | aspect: bathroom, which is connected to (unfriendly, filthy, service, but) within one hop. opinion: filthy, which is connected to (bathroom) within one hop.\\candidate: aspect: service, opinion: ok | aspect: service, opinion: unfriendly | aspect: bathroom, opinion: filthy\\
        \textbf{Output}:\\
        aspect: service, opinion: ok, category: service general, sentiment: negative | aspect: service, opinion: unfriendly, category: service general, sentiment: negative | aspect: bathroom, opinion: filthy, category: ambience general, sentiment: negative \\
        < | end\_of\_sentence | > \\

        \hline
    \end{tabular}
\end{table}
\begin{table}[!htb]
    \centering
    \tiny
    \begin{tabular}{|p{0.45\textwidth}|}
        \hline
        \rule{0pt}{8pt}
        \centerline{\small{\textbf{task: (classification aspect to category)}}}\\
        \hline
        \textbf{Instruction}: \\
        Given a sentence, related dependency relations (will be presented in the form of subgraph) and known aspects (both implicit and explicit) , determine the category related to the each aspects from dependency relation and return pair (aspect, category).\\
        \textbf{Input}: \\
        sentence: service ok but unfriendly , filthy bathroom . \\ subgraph: aspect: service, which is connected to (bathroom, ok) within one hop. | aspect: service, which is connected to (bathroom, ok) within one hop. | aspect: bathroom, which is connected to (unfriendly, filthy, service, but) within one hop.\\candidate aspect: service | service | bathroom\\
        \textbf{Output}:\\
        aspect: service, category: service general | aspect: service, category: service general | aspect: bathroom, category: ambience general \\
        < | end\_of\_sentence | > \\

        \hline
    \end{tabular}
\end{table}
\begin{table}[!htb]
    \centering
    \tiny
    \begin{tabular}{|p{0.45\textwidth}|}
        \hline
        \rule{0pt}{8pt}
        \centerline{\small{\textbf{task: (classification aspect to sentiment)}}}\\
        \hline
        \textbf{Instruction}: \\
        Given a sentence, related dependency relations (will be presented in the form of subgraph) and known aspects (both implicit and explicit) , determine the sentiment related to the each aspects from dependency relation and return pair (aspect, sentiment).\\
        \textbf{Input}: \\
        sentence: service ok but unfriendly , filthy bathroom . \\ subgraph: aspect: service, which is connected to (bathroom, ok) within one hop. | aspect: service, which is connected to (bathroom, ok) within one hop. | aspect: bathroom, which is connected to (unfriendly, filthy, service, but) within one hop.\\candidate aspect: service | service | bathroom\\
        candidates: aspect: service | aspect: service | aspect: bathroom \\
        \textbf{Output}:\
        aspect: service, sentiment: negative | aspect: service, sentiment: negative | aspect: bathroom, sentiment: negative \\
        < | end\_of\_sentence | > \\

        \hline
    \end{tabular}
\end{table}
\begin{table}[!htb]
    \centering
    \tiny
    \begin{tabular}{|p{0.45\textwidth}|}
        \hline
        \rule{0pt}{8pt}
        \centerline{\small{\textbf{task: (classification opinion to category)}}}\\
        \hline
        \textbf{Instruction}: \\
        Given a sentence, related dependency relations (will be presented in the form of subgraph) and known opinions (both implicit and explicit) , determine the category related to the each opinions from dependency relation and return pair (opinion, category).\\
        \textbf{Input}: \\
        sentence: service ok but unfriendly , filthy bathroom . \\ subgraph: opinion: ok, which is connected to (service) within one hop. | opinion: unfriendly, which is connected to (bathroom) within one hop. | opinion: filthy, which is connected to (bathroom) within one hop.\\
        candidate opinion: ok | unfriendly | filthy \\
        \textbf{Output}:\\
        opinion: ok, category: service general | opinion: unfriendly, category: service general | opinion: filthy, category: ambience general \\
        < | end\_of\_sentence | > \\

        \hline
    \end{tabular}
\end{table}
\begin{table}[!htb]
    \centering
    \tiny
    \begin{tabular}{|p{0.45\textwidth}|}
        \hline
        \rule{0pt}{8pt}
        \centerline{\small{\textbf{task: (classification opinion to sentiment)}}}\\
        \hline
        \textbf{Instruction}: \\
        Given a sentence, related dependency relations (will be presented in the form of subgraph) and known opinions (both implicit and explicit) , determine the category related to the each opinions from dependency relation and return pair (opinion, category).\\
        \textbf{Input}: \\
        sentence: service ok but unfriendly , filthy bathroom . \\ subgraph: opinion: ok, which is connected to (service) within one hop. | opinion: unfriendly, which is connected to (bathroom) within one hop. | opinion: filthy, which is connected to (bathroom) within one hop.\\
        candidate opinion: ok | unfriendly | filthy\\
        \textbf{Output}:\\
        opinion: ok, sentiment: negative | opinion: unfriendly, sentiment: negative | opinion: filthy, sentiment: negative \\
        < | end\_of\_sentence | > \\

        \hline

    \end{tabular}
\end{table}
\section{Inference}\label{Appendix B}
In this sention, we will introduce the details of our model during the inference process. The unidirectional dependency of sentiment elements in generative models has been thoroughly discussed in MVP~\citep{mvp} and SLGM~\citep{slgm}. Therefore, in the Global Syntax-guided Extraction, we adopt a bidirectional generation approach for the ⟨a,o⟩ pair, similar to SLGM. Beam search is set to a size of 4.
\end{document}